\documentclass[10pt,twocolumn,letterpaper]{article}

\usepackage[pagenumbers]{cvpr} 

\usepackage[dvipsnames]{xcolor}
\usepackage[utf8]{inputenc} 
\usepackage[T1]{fontenc}    
\usepackage{url}            
\usepackage{booktabs}       
\usepackage{amsfonts}       
\usepackage{amsmath}
\usepackage{nicefrac}       
\usepackage{microtype}      
\usepackage{graphicx}
\usepackage{graphbox}
\usepackage{subcaption}
\usepackage{algorithm}
\usepackage{algorithmic}
\usepackage{bm}
\usepackage[accsupp]{axessibility} 

\newcommand{\TODO}[1]{}
\renewcommand{\TODO}[1]{{\color{cyan} [TODO: {#1}]}}
\newcommand{\WIP}[1]{}
\renewcommand{\WIP}[1]{{\color{magenta} [WIP: {#1}]}}
\definecolor{midblue}{rgb}{0,0.11372549,0.258823529}

\definecolor{cvprblue}{rgb}{0.21,0.49,0.74}
\usepackage[pagebackref,breaklinks,colorlinks,citecolor=cvprblue]{hyperref}

\def\paperID{7769} 
\def\confName{CVPR}
\def\confYear{2024}

\title{Adaptive Random Feature Regularization on Fine-tuning Deep Neural Networks}

\author{%
  Shin'ya Yamaguchi\thanks{Corresponding author. \texttt{shinya.yamaguchi@ntt.com}}\\
  NTT, Kyoto University \\
  \and
  Sekitoshi Kanai \\
NTT \\
  \and
  Kazuki Adachi \\
  NTT\\
  \and
  Daiki Chijiwa \\
  NTT\\
}

\begin{document}

\maketitle

\begin{abstract}
  While fine-tuning is a de facto standard method for training deep neural networks, it still suffers from overfitting when using small target datasets.
  Previous methods improve fine-tuning performance by maintaining knowledge of the source datasets or introducing regularization terms such as contrastive loss.
  However, these methods require auxiliary source information (e.g., source labels or datasets) or heavy additional computations.
  In this paper, we propose a simple method called \textit{adaptive random feature regularization (AdaRand)}.
  AdaRand helps the feature extractors of training models to adaptively change the distribution of feature vectors for downstream classification tasks without auxiliary source information and with reasonable computation costs.
  To this end, AdaRand minimizes the gap between feature vectors and random reference vectors that are sampled from class conditional Gaussian distributions.
  Furthermore, AdaRand dynamically updates the conditional distribution to follow the currently updated feature extractors and balance the distance between classes in feature spaces.
  Our experiments show that AdaRand outperforms the other fine-tuning regularization requiring auxiliary source information and heavy computation costs.
\end{abstract}
\begin{figure*}[t]
    \centering
    \includegraphics[width=0.8\linewidth]{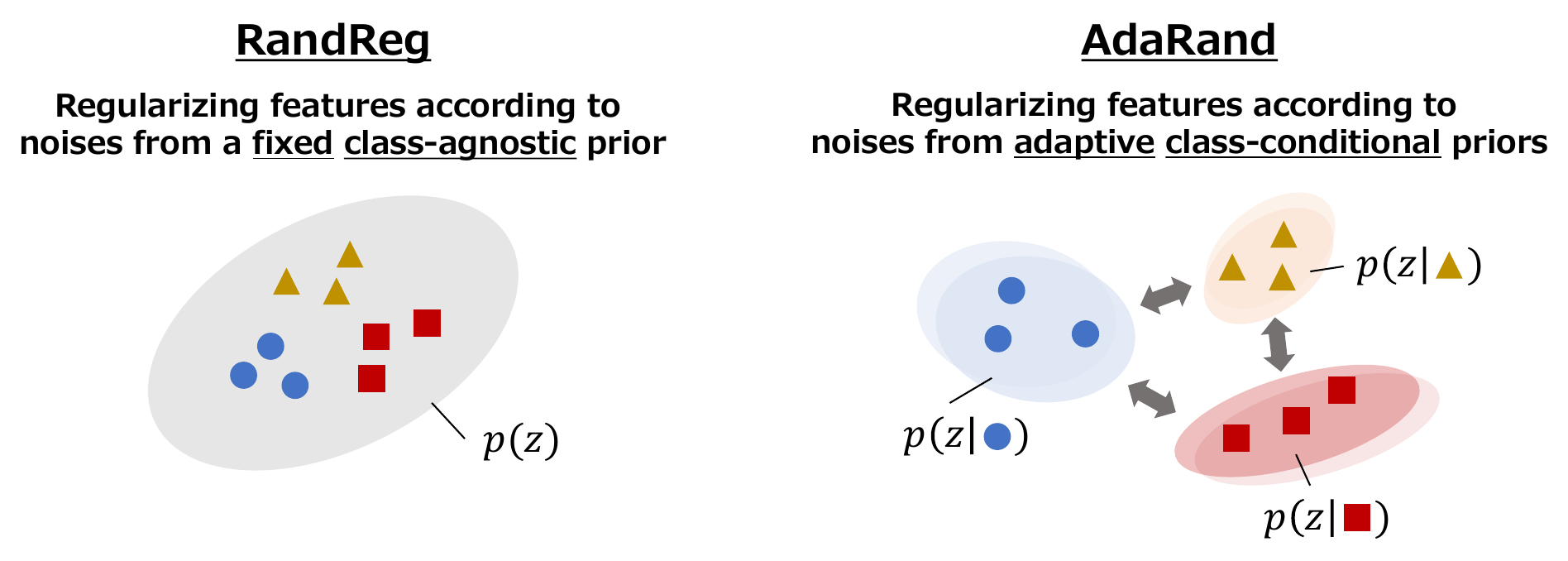}
    \caption{
    Intuitive comparison of RandReg and AdaRand (proposed method).
    RandReg regularizes a feature extractor by minimizing the gap between input features and noises generated from a fixed prior distribution.
    Although RandReg is very simple, it tends to concentrate the features in local regions due to the prior being fixed and class-agnostic, preventing separate classes.
    In contrast, AdaRand adopts class conditional priors and dynamically updates them with the running feature statistics of training models and the maximization distances between each pair of class conditional priors.
    This helps models to obtain more separable features and improve accuracy.
    }
    \label{fig:method}
\end{figure*}
\section{Introduction}\label{sec:intro}
Fine-tuning is a standard technique for training deep neural network models.
In fine-tuning, we pre-train a model on large-scale source tasks (e.g., ImageNet~\cite{russakovsky_imagenet} and WebImageText for CLIP~\cite{Radford_ICML21_CLIP}) before training it on target tasks.
Fine-tuning can improve the performance and efficiency of training on the target task~\cite{he_ICCV19_rethinking_finetuning}.
However, fine-tuning deep neural networks still suffers from overfitting when small target datasets are used~\cite{xuhong_ICML18_l2sp}.

To alleviate the overfitting, previous studies have proposed regularization terms so that models maintain source knowledge~\cite{xuhong_ICML18_l2sp,li_ICLR19_delta,you_NeurIPS20_co_tuning,Liu_NeurIPS2022_improved_finetuning_by_source_data}.
The early methods simply minimize the gap between source and target models on the model parameters~\cite{xuhong_ICML18_l2sp} or the intermediate outputs~\cite{li_ICLR19_delta}.
The recent methods have evolved to utilize auxiliary source information such as source classification labels~\cite{you_NeurIPS20_co_tuning} and source datasets~\cite{Liu_NeurIPS2022_improved_finetuning_by_source_data} to improve the performance.
However, the source information is not always available because pre-training methods are not limited to supervised classification and source datasets are often private.
For instance, multimodal pre-training by CLIP~\cite{Radford_ICML21_CLIP} provides only pre-trained parameters but not the source dataset (WebImageText).
Thus, we cannot directly apply the regularization methods using the auxiliary source information to CLIP pre-trained models.

An alternative regularization strategy that does not require source information is to modify feature vectors or task heads to be desirable for solving downstream classification tasks~\cite{Hariharan_ICCV17_feature_penalty,Zhong_CVPR20_RandReg,Zhang_NeurIPS21_coretuning, Zhou_ICCV23_drtune}.
These methods penalize the feature extractors of fine-tuning models by \(\ell_1/\ell_2\) loss for selecting features to train effectively~\cite{Hariharan_ICCV17_feature_penalty,Zhong_CVPR20_RandReg} or contrastive loss for obtaining instance discriminative features~\cite{Zhang_NeurIPS21_coretuning}.
More recently, \citet{Zhou_ICCV23_drtune} have shown a regularization method that enforces linear classification head parameters to classify the features extracted from fixed source models.
However, this method requires additional memory to store pre-trained features and compute their penalty term. This increases the memory and computation costs of fine-tuning and the complexity of implementation.

To achieve high performance without auxiliary source information and non-negligible computation costs, we spotlight the method proposed by~\citet{Zhong_CVPR20_RandReg}, which we call random feature regularization (RandReg).
RandReg penalizes the feature extractor by the \(\ell_2\) distance between feature vectors and random reference vectors.
The reference vectors are drawn from a class-agnostic prior distribution that is independent of the target task, e.g., uniform distribution.
By perturbing features with randomness in addition to penalizing the feature norms, RandReg boosts fine-tuning performance without auxiliary source information and heavy additional computation costs.

Although RandReg boosts the performance, we empirically found that (a) the performance gain of RandReg depends on the choice of the prior distribution because the feature norms and scales of pre-trained models vary widely depending on the pre-training methods (Table~\ref{tb:prior_multi_model}), 
(b) in some cases, RandReg is unexpectedly inferior to the \(\ell_2\) feature regularization without randomness,
(c) RandReg has an unexpected effect of reducing the feature norm and entropy due to the single class-agnostic prior, which leads to limiting gradients of cross-entropy loss and mutual information between features and target labels.
Therefore, na\"ively introducing RandReg with a simple prior may limit the performance of target models.

To address the challenges of RandReg, we propose \textit{adaptive random feature regularization} (AdaRand), which extends RandReg to be effective for arbitrary pre-training methods including self-supervised learning and CLIP.
AdaRand uses a parametric class conditional Gaussian prior that is dynamically updated during fine-tuning instead of a fixed class-agnostic prior.
By initializing the prior distribution with the statistics of feature vectors computed on pre-trained models for each target class, AdaRand performs regularization stably without suffering from the differences in features due to the choice of pre-training methods.
Whereas RandReg causes small feature norms and entropy, AdaRand prevents them by dynamically updating the prior parameters of each class according to the fine-tuning process.
The objective function consists of (i) fitting the mean parameters to the class-wise running mean of feature vectors during fine-tuning and (ii) penalizing them so that they are not similar to any other class.
That is, the prior distributions are moved toward the distribution of the current feature vectors while maintaining a margin between classes. 
This improves the mutual information between features and target labels, resulting in separable clusters of feature vectors that are suitable for the target classification task (Figure~\ref{fig:method}).

We conduct experiments to evaluate AdaRand with various (classification, self-supervised, and CLIP) pre-training methods on multiple datasets.
The experiments show that AdaRand outperforms RandReg and existing fine-tuning methods depending on auxiliary source information and non-negligible computation costs, even though AdaRand does not require either.

\section{Related Work}\label{sec:relatedwork}
Many regularization methods for fine-tuning deep neural networks are based on the assumption that maintaining source knowledge is beneficial for solving target tasks.
According to this assumption, \citet{xuhong_ICML18_l2sp} have presented a simple regularization called L2SP, which minimizes the parameters between source and target models during fine-tuning.
A subsequent study~\cite{li_ICLR19_delta} has shown that learning to maximize the similarity between the output feature vectors of source and target models can outperform L2SP.
To prevent catastrophic forgetting and negative transfer, batch spectral shrinking (BSS, \cite{Chen_NeurIPS19_BSS}) penalizes smaller singular values of the batch feature matrices.
Although these methods are simple and flexible for arbitrary pre-training methods, the performance improvements are limited.
To achieve more practical performance, Co-tuning~\cite{you_NeurIPS20_co_tuning} leverages source knowledge contained in the source task-specific layers on the head of pre-trained models, which are often discarded during fine-tuning.
Specifically, in addition to target tasks, Co-tuning simultaneously solves a pseudo-source task that is defined by soft-source labels corresponding to each target label.
BTfW~\cite{Ge_CVPR17_BTfW} and UOT~\cite{Liu_NeurIPS22_UOT} search target-related subsets of the source dataset through the selection algorithm and train a model on both the target and target-related source subset to directly transfer source knowledge.
By leveraging auxiliary source information (i.e., source class labels and datasets), Co-tuning and UOT have achieved impressive fine-tuning performance on target tasks.
However, since recent powerful pre-training models such as CLIP~\cite{Radford_ICML21_CLIP} are often not trained on classification or publicly available datasets, we cannot apply the previous methods to them.

On the other hand, there are regularization methods that refine parameters or features without explicitly maintaining source knowledge.
\citet{Takada_NeurIPS20_transfer_lasso} have shown that the \(\ell_1\) regularization on training parameters helps to select the parameters to be updated and improve fine-tuning performance.
\citet{Hariharan_ICCV17_feature_penalty} have proposed the \(\ell_1\)/\(\ell_2\) regularization methods on feature vectors called feature norm penalty (FNP), which restrict feature activation to extract only useful information with limited volume target data.
Subsequently, \citet{Zhong_CVPR20_RandReg} have proposed RandReg, which minimizes the gap between feature vectors and the random reference vectors from a uniform prior distribution.
RandReg can be regarded as an advanced method of FNP because the definition of RandReg is decomposed by the \(\ell_1\)/\(\ell_2\) regularization term and perturbation term, which is designed to help models not to be trapped in local minima~\cite{Zhong_CVPR20_RandReg}.
Our work is positioned as one of the regularization methods without the assumption of auxiliary source information and improves RandReg by introducing class conditional Gaussian priors that are dynamically updated to prevent the small norm and low entropy features.

The assumption of conditional Gaussian (mixture) distributions over the feature spaces is often used in the context of adversarial robustness~\cite{Pang_ICML18_MaxMahalanobis_robust,Wan_TPAMI22_GMM_robust}.
In contrast, our method aims to improve the performance of fine-tuning by regularizing the feature extractor with the reference vectors sampled from the conditional distributions.

\section{Preliminary}\label{sec:preliminary}
\subsection{Problem Setting}
In this paper, we consider a standard problem of fine-tuning deep neural networks on a \(K\) class classification task. 
We train a neural network model \(f_{\theta} :\mathcal{X} \to \mathcal{Y}\) on a labeled target dataset \(\mathcal{D}=\{(x^i,y^i) \in \mathcal{X}\times\mathcal{Y}\}^{N}_{i=1}\), where \(\mathcal{X}\) and \(\mathcal{Y}\) are the input and output label spaces, respectively.
\(f_{\theta}\) is defined by a composition of a feature extractor \(g_{\phi}: \mathcal{X}\to\mathbb{R}^d\) and a weight matrix for linear classification \(W \in\mathbb{R}^{d\times K}\), i.e., \(f_{\theta} = W^\top g_\phi\) and \(\theta = [\phi, W]\).
Here, \(\theta\) is initialized by \(\theta_\mathrm{s}=[\phi_\mathrm{s},W_\mathrm{s}]\), which is pre-trained on large-scale source datasets through arbitrary pre-training methods such as supervised training~\cite{yosinski_NeurIPS14_transferable}, self-supervised contrastive training~\cite{Chen_ICML20_SimCLR, He_CVPR20_MoCo}, and multimodal training~\cite{Radford_ICML21_CLIP}.

\subsection{Random Feature Regularization}
We recall the principle of random feature regularization (RandReg)~\cite{Zhong_CVPR20_RandReg}.
RandReg is a regularization method for fine-tuning deep neural networks that uses a prior distribution \(p(z)\) of the random reference vector \(z\in\mathbb{R}^d\), i.e., \(z\sim p(z)\).
The objective function is defined as follows:
\begin{eqnarray}
     &\min\limits_{\theta=[\phi,W]}   \mathcal{L}_\mathrm{cls}(\theta)+\lambda\mathcal{L}_\mathrm{reg}(\phi),\label{eq:obj_randreg}\\
    &\mathcal{L}_\mathrm{cls}(\theta) = {\mathbb{E}_{(x,y)\in \mathcal{D}}}~\ell_\mathrm{CE}(f_\theta(x),y),\label{eq:cls_loss}\\
    &\mathcal{L}_\mathrm{reg}(\phi) = {\mathbb{E}_{x\in \mathcal{D}}}~\|g_\phi(x)-z\|^2_2,\label{eq:randreg_loss}
\end{eqnarray}
where \(\ell_\mathrm{CE}\) is cross-entropy loss.
Intuitively, by minimizing the gap between \(g_\phi(x)\) and \(z\), \(\mathcal{L}_\mathrm{reg}(\phi)\) makes the feature extractor \(g_\phi\) forms the output feature vectors according to the prior distribution \(p(z)\).
The original paper~\cite{Zhong_CVPR20_RandReg} explains that RandReg improves classifiers because the reference vector \(\bm{z}\) enlarges the variance of the gradient and prevents the model from overfitting.
However, we found that the performance gain of RandReg largely depends on the combinations of pre-training methods and priors, and the na\"ive randomness by RandReg is not effective in some cases.

\begin{table*}[t]
    \centering
    \caption{
    Analysis of random feature regularization (RandReg~\cite{Zhong_CVPR20_RandReg}) in top-1 test accuracy on various pre-training methods (Cars, ResNet-50). We also report the statistics of target data computed on a pre-trained model: the averaged feature norms (\(\|g_{\phi_\mathrm{s}}(x)\|\)) and the dimension-wise averaged mean and variance (\((\bar{\mu}_\mathrm{s}=\sum^d_{i=1} \mu_\mathrm{s}[i],\bar{\sigma}^2_\mathrm{s}=\sum^d_{i=1} \sigma^2_\mathrm{s}[i]))\)). The performance depends on the combinations of pre-training methods and prior. Moreover, there is a case that the accuracy of RandReg degrades from simple \(\ell_2\) feature regularization (i.e., FNP~\cite{Hariharan_ICCV17_feature_penalty}).
    }
    \label{tb:prior_multi_model}
            \resizebox{1.0\linewidth}{!}{
        \begin{tabular}{l|cc|ccccc}\toprule
            Pre-trained Method  & \(\|g_{\phi_\mathrm{s}}(x)\|^2_2\) & \((\bar{\mu}_\mathrm{s},\bar{\sigma}^2_\mathrm{s})\) & Fine-tuning & FNP~\cite{Hariharan_ICCV17_feature_penalty} & RandReg-\(U(0,1)\) & RandReg-\(\mathcal{N}(0,1)\) & RandReg-\(\mathcal{N}(\mu_\mathrm{s},\sigma^2_\mathrm{s})\) \\
          \midrule
            ImageNet Classification  & 19.58  & (4.18\(\times10^{-1}\), 2.69\(\times10^{-1}\)) &  89.14\(^{\pm\text{.42}}\) & 90.27\(^{\pm\text{.10}}\)&90.59\(^{\pm\text{.24}}\) &  90.42\(^{\pm\text{.12}}\) & \textbf{90.61}\(^{\pm\textbf{.24}}\) \\
            ImageNet SimCLR~\cite{Chen_ICML20_SimCLR}  & 1.34 & (2.60\(\times10^{-2}\), 3.97\(\times10^{-2}\)) &  83.73\(^{\pm\text{.73}}\) &\textbf{84.53}\(^{\pm\textbf{.32}}\) &84.08\(^{\pm\text{.21}}\) & 84.03\(^{\pm\text{.07}}\) & 83.91\(^{\pm\text{.04}}\)\\
            ImageNet Barlow Twins~\cite{Zbontar_ICML21_barlowtwins} & 3.67 & (5.53\(\times10^{-2}\), 6.94\(\times10^{-2}\)) &  86.98\(^{\pm\text{.16}}\) & 87.44\(^{\pm\text{.15}}\) &  87.24\(^{\pm\text{.30}}\) & \textbf{87.74}\(^{\pm\textbf{.33}}\) & 87.65\(^{\pm\text{.48}}\) \\
            CLIP~\cite{Radford_ICML21_CLIP} & 11.70 & (5.92\(\times10^{-4}\), 3.12\(\times10^{-2}\)) &  88.72\(^{\pm\text{.24}}\) &  89.96\(^{\pm\text{.05}}\) & 90.19\(^{\pm\text{.40}}\)& 90.59\(^{\pm\text{.24}}\) & \textbf{90.78}\(^{\pm\textbf{.07}}\) \\
            \bottomrule
        \end{tabular}
        }
\end{table*}

\section{Observations of RandReg}\label{sec:analisys}
In this section, we analyze RandReg through preliminary experiments from the perspective of pre-training methods and priors.
We evaluated RandReg with pre-trained ResNet-50~\cite{he_resnet} models. We provide more detailed training settings in Sec.~\ref{sec:ex_setting}.
In summary, we obtained three observations.
\begin{itemize}
    \item Effective prior distribution depends on pre-training methods and the features of the pre-trained models.
    \item RandReg underperforms the \(\ell_2\) feature regularization without randomness (i.e., FNP~\cite{Hariharan_ICCV17_feature_penalty}), indicating that there are cases where the na\"ive randomness is not effective.
    \item RandReg makes training models generate features with small norms and less diversity.
\end{itemize}
Consequently, based on these observations, we discuss the challenges of RandReg in terms of fixed prior distributions and features with small norms and less diversity.

\subsection{Effects of Prior Distribution}\label{sec:pre_prior_choice}
We examine the performance when varying prior distributions in RandReg.
We tried three priors for training target models: \textbf{uniform distribution} \(U(0,1)\), \textbf{standard Gaussian distribution} \(\mathcal{N}(0,I)\), and \textbf{Gaussian distribution with pre-computed statistics} \(\mathcal{N}(\mu_\mathrm{s}, \sigma^2_\mathrm{s}I)\), which is parameterized by feature mean \(\mu_\mathrm{s} \in \mathbb{R}^d\) and variance \(\sigma_\mathrm{s}^2 \in \mathbb{R}^d\) computed with pre-trained weights on the target dataset.
To assess the importance of the randomness in RandReg, we also tested the \(\ell_2\) feature norm regularization without randomness (FNP, \cite{Hariharan_ICCV17_feature_penalty}), which is equivalent to the case where \(z\) is 0 in Eq.~(\ref{eq:randreg_loss}).
We evaluated these priors on multiple pre-trained models with four pre-training methods including classification~\cite{agrawal_ECCV14_analyzing_finetuning}, SimCLR~\cite{Chen_ICML20_SimCLR}, Barlow Twins~\cite{Zbontar_ICML21_barlowtwins}, and CLIP~\cite{Radford_ICML21_CLIP}.
Table~\ref{tb:prior_multi_model} shows the top-1 test accuracy on the Cars~\cite{krause_3DRR2013_stanford_cars} dataset.
We observed that RandReg improves the fine-tuning baselines for all pre-trained models, but the best prior depends on pre-trained methods.
Table~\ref{tb:prior_multi_model} also lists the averaged feature norms (\(\|g_{\phi_\mathrm{s}}(x)\|\)) and the dimension-wise averaged mean and variance (\(\bar{\mu}_\mathrm{s}=\sum^d_{i=1} \mu_\mathrm{s}[i],\bar{\sigma}^2_\mathrm{s}=\sum^d_{i=1} \sigma^2_\mathrm{s}[i]\)) that are computed on each pre-trained feature extractor by forwarding target data before fine-tuning.
Uniform and standard Gaussian distributions are effective when the feature scale is small (e.g., Barlow Twins), but when the scale is large (e.g., ImageNet Classification), Gaussian distributions with pre-computed statistics, i.e., \(\mathcal{N}(\mu_\mathrm{s},\sigma_\mathrm{s}^2I)\), are the best.
This implies that considering a gap between the feature distributions of pre-trained models and prior distributions is important in selecting priors.
However, RandReg is inferior to FNP in the case of SimCLR, where the feature norm and scale are quite small, even when we use pre-computed statistics.
This means that the na\"ive randomness introduced by RandReg is not effective for fine-tuning in some cases.

\subsection{Effects on Feature Norm and Diversity}\label{sec:pre_feature}
\begin{figure*}[t]
    \centering
    \begin{minipage}{0.24\linewidth}
    \centering
        \includegraphics[width=\columnwidth]{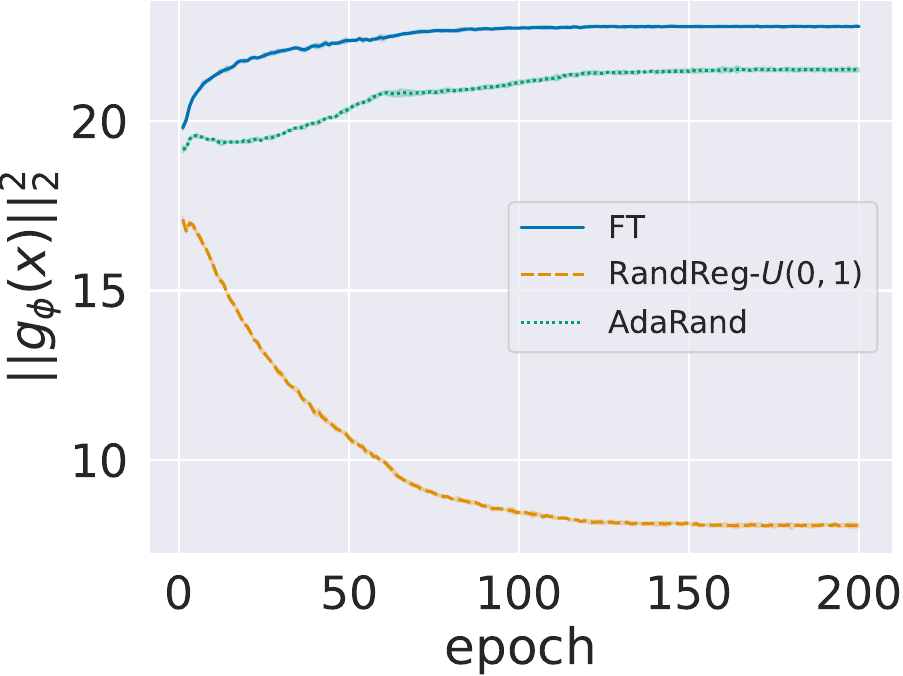}
    \subcaption{Feature Norm}\label{fig:feat_norm}
    \end{minipage}
    \begin{minipage}{0.24\linewidth}
    \centering
        \includegraphics[width=\columnwidth]{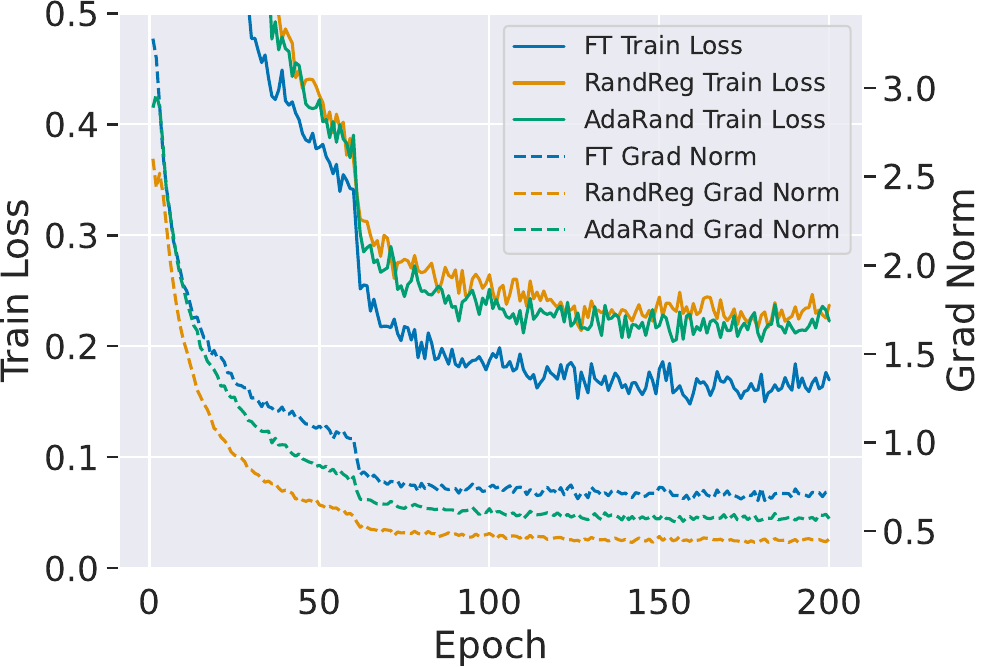}
    \subcaption{Gradient Norm}\label{fig:grad_norm}
    \end{minipage}
    \begin{minipage}{0.24\linewidth}
    \centering
        \includegraphics[width=1.0\columnwidth]{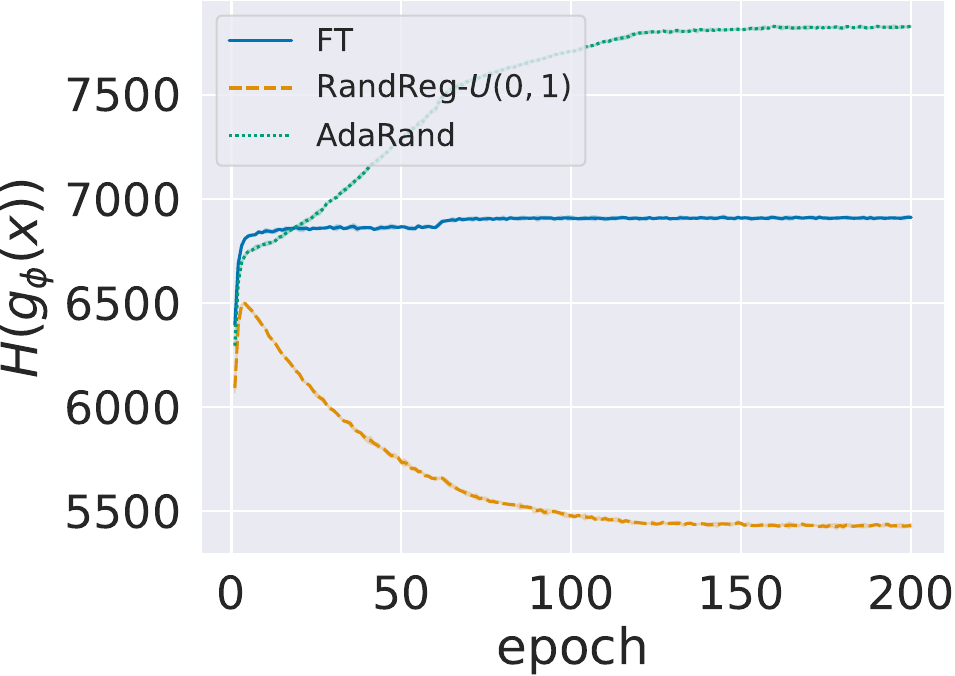}
    \subcaption{Feature Entropy}\label{fig:feat_entropy}
    \end{minipage}
    \begin{minipage}{0.24\linewidth}
    \centering
        \includegraphics[width=\columnwidth]{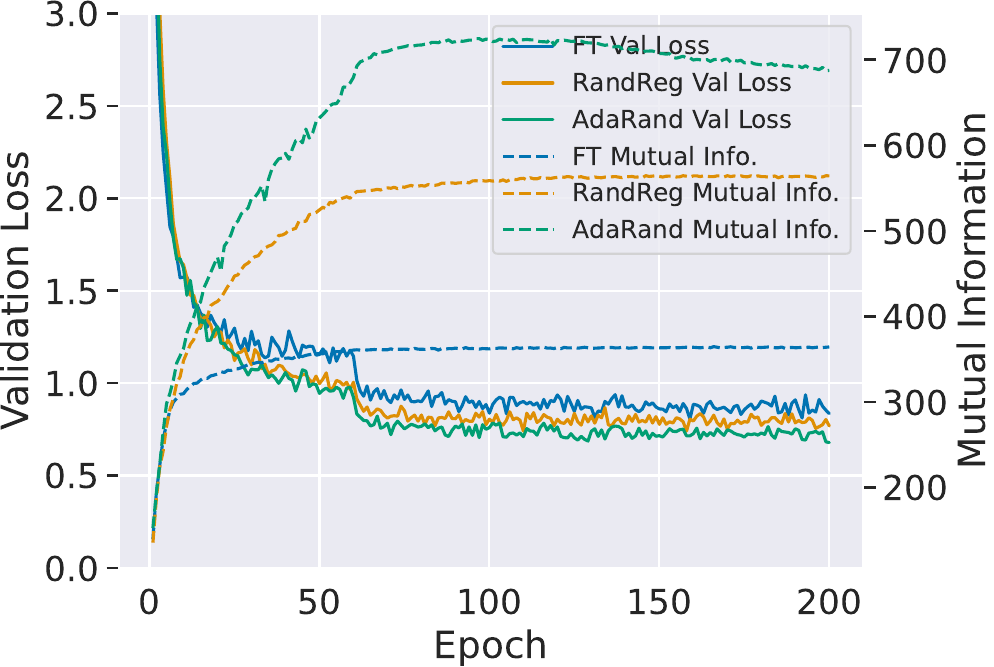}
    \subcaption{Mutual Information}\label{fig:feat_mi}
    \end{minipage}
    \vspace{-3mm}
    \caption{
    Statistics on feature vectors during training (Cars, ResNet-50 pre-trained with ImageNet classification). RandReg tends to decrease the feature norm \(||g_\phi(x)||^2_2\) and entropy \(H(g_\phi(x))\). This implicitly limits the gradient norm of cross-entropy loss \(\|\nabla_W \ell_\mathrm{CE}(f_\theta(x), y)\|^2_2\) and the discriminability of features represented by mutual information \(I(g_\phi(x); y)\).
    In contrast, our AdaRand prevents the models from decreasing \(||g_\phi(x)||^2_2\) and \(H(g_\phi(x))\), and thus achieves larger \(\|\nabla_W \ell_\mathrm{CE}(f_\theta(x), y)\|^2_2\) and \(I(g_\phi(x); y)\) than RandReg.
    }\label{fig:norm_entropy}
    \vspace{-3mm}
\end{figure*}

Next, we investigate the effects on features caused by RandReg.
Here, we focus on the feature norm and diversity. 
This is because the regularization term of RandReg in Eq.~(\ref{eq:randreg_loss}) directly affects the feature norm \(\|g_\phi(x)\|^2_2\) by the squared loss, and it can restrict the diversity of the features by the prior \(p(z)\).
We interpret the feature diversity as the feature entropy \(H(g_\phi(x))\), and estimate \(H(g_\phi(x))\) by the differential entropy estimator with the assumption that the probabilistic density of \(g_\phi(x)\) is constant in an \(\epsilon\)-ball around a feature \(g_\phi(x^i)\) for a randomly sampled \(x^i\)~\cite{Faivishevsky_NeurIPS08_entropy_estimator}:
\begin{eqnarray}
    {H}(g_\phi(x)) \approx \frac{d}{N(N-1)}\sum_{i\neq j}{\log \|g_\phi(x^i) - g_\phi(x^j)\|^2_2},
\end{eqnarray}
where \(d\) is the dimension of a feature vector and \(N\) is the dataset size.
Figure~\ref{fig:norm_entropy} plots the feature norm \(\|g_\phi(x)\|^2_2\) and entropy \(H(g_\phi(x))\) for each epoch in training; we calculated \(\|g_\phi(x)\|^2_2\) and \(H(g_\phi(x))\) on training samples.
While fine-tuning slightly increases \(\|g_\phi(x)\|^2_2\) and \(H(g_\phi(x))\), RandReg gradually decreases both of them, which means RandReg produces small feature vectors with less diversity.

\subsection{Challenges of RandReg}\label{sec:pre_challenges}
\paragraph{Fixed prior distribution.}
From the results in Sec.~\ref{sec:pre_prior_choice}, although RandReg can improve fine-tuning regardless of pre-training methods, we should select an effective prior distribution to adjust the reference vectors to the pre-trained feature extractor and obtain the best performance.
This is challenging in practice because it is not obvious which prior distribution is the best choice for the pre-training method and the manual search is costly; na\"ively using pre-computed statistics \(\mu_\mathrm{s}\) and \(\sigma^2_\mathrm{s}\) did not always achieve the best performance in Table~\ref{tb:prior_multi_model}.
If the prior is not effective, there is a risk that the randomness will not work effectively  as in the case of SimCLR in Table~\ref{tb:prior_multi_model}.
Thus, we need to efficiently search for appropriate prior distributions in fine-tuning.
\vspace{-3mm}
\paragraph{Small norms and less diversity.}
The experiments in Sec.~\ref{sec:pre_feature} show that RandReg degenerates the feature norm \(\|g_\phi(x)\|^2_2\) and entropy \(H(g_\phi(x))\).
This poses potential challenges for fine-tuning.
First, the small feature norm may vanish the gradient of target loss functions.
In a classification task, the gradient of the cross-entropy loss with respect to \(W\) of the classifier is formulated as follows~\cite{Hariharan_ICCV17_feature_penalty}.
\begin{eqnarray}
    \|\nabla_W \ell_\mathrm{CE}(f_\theta(x), y)\|^2_2 = \sum^K_{k=1} \left(f_\theta(x)[k] - \delta_{yk}\right)^2\|g_\phi(x)\|^2_2,\label{eq:ce_grad}
\end{eqnarray}
where \(f_\theta(x)[k]= W^\top g_\phi(x)[k]\) is the output of the classifier for the \(k\)-th class that is normalized by softmax function, and \(\delta_{yk}\) is \(1\) if \(y=k\) otherwise \(0\).
Since Eq.~(\ref{eq:ce_grad}) contains the product of \(\|g_\phi(x)\|^2_2\), degenerating \(\|g_\phi(x)\|^2_2\) leads to vanishing \(\|\nabla_W \ell_\mathrm{CE}\|^2_2\), resulting in stagnation of fine-tuning.
Second, the low entropy limits the model's ability to learn effective feature representations for solving the target task in terms of mutual information, which is strongly related to model performance~\cite{Hjelm_ICLR19_deepinfomax}.
Mutual information \(I(g_\phi(x);y)\) measures the amount of shared information (i.e., correlation) between the feature vector \(g_\phi(x)\) and label \(y\) as defined by
\begin{eqnarray}
     I(g_\phi(x);y) = H(g_\phi(x)) - H(g_\phi(x)|y).\label{eq:mi}
\end{eqnarray}
The first term on the right-hand side \(H(g_\phi(x))\) can be interpreted as the ``diversity'' of features and the second term \(- H(g_\phi(x)|y)\) as the ``tightness'' for each class~\cite{Boudiaf_ECCV20_mutual_infor_metric,Zhang_ICLR23_ordinal_entropy_regression}.
In this sense, RandReg limits the mutual information of the feature representations by decreasing the diversity term \(H(g_\phi(x))\). 
Although RandReg can increase the mutual information than the baseline as shown in Fig.~\ref{fig:feat_mi}, there is a potential to learn a more effective feature representation if we resolve the decrease of \(H(g_\phi(x))\).

\begin{algorithm}[t]
    \caption{AdaRand}\label{alg:adarand}
    \begin{algorithmic}[1]
    {\small
        \REQUIRE{Training dataset \(\mathcal{D}\), target model \(f_\theta\), training batchsize \(B\), step size \(\eta\) and \(\xi\), trade-off parameter \(\lambda\)
        \ENSURE{Trained classifier \(f_\theta\)}}
        \STATE{Initialize \(\boldsymbol{\mu}\) and \(\boldsymbol{\sigma}^2\) for conditional priors by Eq.~(\ref{eq:init_mu})~and~(\ref{eq:init_sigma})}.
        \STATE{\(\bar{\boldsymbol{\mu}} \leftarrow \boldsymbol{\mu}\)}
        \WHILE{not converged}
        \STATE{\(\{(x^i,y^i)\}^B_{i=1}\sim \mathcal{D}\)}
        \STATE{\(\{z^i | z^i \sim \mathcal{N}(\mu_{y^i},\sigma^2_{y^i})\}^B_{i=1}\)}
        \STATE{// Updating \(f_\theta = W^\top g_\phi\)}
        \STATE{\(\theta \leftarrow \theta - \frac{\eta}{B}\sum_{i=1}^{B}\!\!\nabla_\theta(\ell_\mathrm{CE}(f_\theta(x^i),y^i)\!+\!\lambda \|g_\phi(x^i)-z^i\|^2_2 )\)}
        \STATE{// Updating conditional priors}
        \STATE{\(\bar{\boldsymbol{\mu}} \leftarrow \operatorname{ema\_update}(\{(g_\phi(x^i),y^i)\}^B_{i=1})\) // Eq.~(\ref{eq:ema_update}),(\ref{eq:batch_mean})}
        \STATE{\(\boldsymbol{\mu} \leftarrow \boldsymbol{\mu} - \xi\nabla_{\boldsymbol{\mu}}(\ell_\mathrm{intra}(\boldsymbol{\mu},\bar{\boldsymbol{\mu}})\!+\! \ell_\mathrm{inter}(\boldsymbol{\mu}))\) // Eq.~(\ref{eq:ada_objective})-(\ref{eq:loss_inter})}
        \ENDWHILE
        }
    \end{algorithmic}
\end{algorithm}
\vspace{-3mm}

\section{Proposed Method}\label{sec:method}
We propose adaptive random feature regularization (AdaRand) by extending RandReg in terms of prior distributions.
AdaRand adopts adaptive class conditional priors, which automatically search for prior parameters and prevent feature vectors from degenerating the norm and entropy (Fig.~\ref{fig:method}).
The overall procedure of AdaRand is described in Algorithm~\ref{alg:adarand}.
Before fine-tuning, we initialize the conditional priors by computing the mean and variance of feature vectors of target data on pre-trained models.
During fine-tuning, we regularize target models by the reference vectors from the conditional priors and then update the mean parameters of the prior by approaching them to the running feature mean on training models (\(\ell_\mathrm{intra}\)) and maximizing the distances between classes (\(\ell_\mathrm{inter}\)).

\subsection{Conditional Prior}
\paragraph{Overview.}
To overcome the challenge of the less diverse feature vectors in RandReg, we introduce conditional prior distributions for generating the reference vectors.
As illustrated in Fig.~\ref{fig:method}, RandReg uses class-agnostic prior such as a uniform distribution.
This prevents models from naturally enlarging the marginal entropy \(H(g_\phi(x))\) to classify labels of input data.
In order to achieve high performance in target classification tasks, the clusters of feature vectors should be separated for each class.
In this sense, the reference vector for the regularization term should be conditioned by class labels of input data. 
Then, the prior of AdaRand can be regarded as the marginal distribution over class labels.
\vspace{-3mm}
\paragraph{Definition.}
The prior distribution is defined as follows.
\begin{eqnarray}
    p(z) = \sum^{K}_{k=1}{p(z|y_k)p(y_k)},\label{eq:adarand_prior}
\end{eqnarray}
where \(p(z|y_k) = \mathcal{N}(\mu_{k},\sigma^2_{k}I)\).
We formalize \(p(z)\) as a mixture of diagonal Gaussian distributions because the feature vectors of softmax neural classifiers are known to follow class conditional Gaussian distributions~\cite{Bishop_95_Neural,Lee_NeurIPS18_simple_ood_gmm}.
When sampling the reference vectors, we generate a random reference vector \(z^i\) from the class label \(y^i\) of the input \(x^i\), i.e.,
\begin{eqnarray}
    z^i \sim \mathcal{N}(\mu_{y^i},\sigma^2_{y^i}I).\label{eq:adarand_prior_sampling}
\end{eqnarray}
Using these conditional reference vectors in Eq.~(\ref{eq:randreg_loss}), we can expect that the regularization effect of random noise can be enjoyed while avoiding the decrease of \(H(g_\phi(x))\) due to class-agnostic prior distributions.
\vspace{-3mm}
\paragraph{Initialization.}
For the prior distribution, we initialize \(\mu_k\) and \(\sigma^2_k\) to the statistics of the pre-trained model and update them adaptively.
As discussed in Sec.~\ref{sec:pre_challenges}, the hyperparameter searches for the initial parameters are challenging.
Therefore, we initialize \(\mu_k\) and \(\sigma^2_k\) with the class-wise mean and variance of feature vectors computed on target data through pre-trained models as 
\begin{eqnarray}
    &\mu_k = \frac{1}{N_k}\sum^{N_k}_{i=1} g_{\phi_\mathrm{s}}(x^i),\label{eq:init_mu}\\
    &\sigma^2_k = \frac{1}{N_k}\sum^{N_k}_{i=1} (g_{\phi_\mathrm{s}}(x^i)-\mu_k)^2,\label{eq:init_sigma}
\end{eqnarray}
where \(N_k\) is the number of samples labeled as a class \(k\) and \(g_{\phi_\mathrm{s}}\) is the pre-trained feature extractor before fine-tuning.
For simple notation in the latter sections, we denote the sets of mean and variance parameters as \(\boldsymbol{\mu}=\{\mu_k\}^K_{k=1}\) and \(\boldsymbol{\sigma}^2=\{\sigma^2_k\}^K_{k=1}\).

\subsection{Adaptive Prior Update}
\paragraph{Overview.}
During fine-tuning, we adaptively optimize the prior parameter set \(\boldsymbol{\mu}\) to adjust the reference vectors according to the training progress of the feature extractors.
This addresses the challenges of fixed prior distributions and small feature norms of RandReg.
That is, we aim to avoid producing small feature norms without manually searching for prior parameters.
Furthermore, we penalize \(\boldsymbol{\mu}\) so that each \(\mu_k\) is independent to the other class parameters \(\{\mu_l\}_{l\neq k}\).
This facilitates models to form tight class conditional feature clusters, corresponding to maximizing the tightness term \(-H(g_\phi(x)|y)\) of the mutual information given by Eq.~(\ref{eq:mi}).

\paragraph{Objective Function.}
The objective function for updating \(\boldsymbol{\mu}\) is defined as follows.
\begin{eqnarray}
    &\mathcal{L}_\mathrm{ada} = \ell_\mathrm{intra}(\boldsymbol{\mu}, \bar{\boldsymbol{\mu}}) + \ell_\mathrm{inter}(\boldsymbol{\mu}),\label{eq:ada_objective}\\
    &\ell_\mathrm{intra}(\boldsymbol{\mu}, \bar{\boldsymbol{\mu}}) = \frac{1}{K}\sum^K_{k=1} D(\mu_k, \bar{\mu}_k)\label{eq:loss_intra},\\
    &\ell_\mathrm{inter}(\boldsymbol{\mu}) = - \frac{1}{K(K-1)}\sum^K_{k=1}\sum^K_{l\neq k} D(\mu_k, \mu_l),\label{eq:loss_inter}
\end{eqnarray}
where, \(\bar{\boldsymbol{\mu}}\) is the running mean vectors of \(g_\phi(x)\) in training, \(D(\cdot)\) is a distance function such as cosine distance, i.e., \(1-\frac{\mu_k\cdot \bar{\mu}_k}{\|\mu_k\|_2\|\bar{\mu}_k\|_2}\).
We update \(\bar{\boldsymbol{\mu}}\) by exponential moving average (EMA) with training samples in batch for each step as
\begin{eqnarray}
    &\bar{\mu}_k \leftarrow \alpha \bar{\mu}_k + (1-\alpha) \hat{\mu}_k,\label{eq:ema_update}\\
    &\hat{\mu}_k = \frac{1}{B_k}\sum^{B_k}_{i=1} g_\phi(x^i),\label{eq:batch_mean}
\end{eqnarray}
where \(\alpha\) is a parameter for controlling the decay of the past information.
We fix \(\alpha=0.5\) throughout this paper; we evaluate the effect of \(\alpha\) in the supplement.
If there are no samples belonging to a class \(k\) in the batch, the EMA update will be skipped.
For \(D(\cdot)\), we used the cosine distance in our experiments because it achieved the best performance empirically.
Intuitively, \(\ell_\mathrm{intra}\) makes \(\mu_k\) approach the feature region that the model is currently learning by fine-tuning, and encourages the model to focus on learning the current region through the reference random vectors.
Meanwhile, $\ell_\mathrm{inter}$ corresponds to maximizing $H(g_\phi(x))$ because the $k$-th class mean parameter $\mu_k$ is penalized for moving away from other class parameters $\{\mu_l\}_{l\neq k}$. Then, the class-wise reference vectors help to gather features for each class (i.e., maximizing $- H(g_\phi(x)|y)$).\looseness-1

\section{Experiments}\label{sec:experiments}
We evaluate AdaRand on the combinations of six visual classification tasks, four pre-training methods, and three neural network architectures.
Furthermore, we conduct qualitative and quantitative experiments to assess the feature space through PCA visualization, feature norms, loss gradients, and mutual information.
We also show the other experiments, including fine-tuning CLIP on ImageNet, in the supplementary.

\subsection{Settings}\label{sec:ex_setting}
\paragraph{Baselines w/o source information.}
We compare AdaRand with the following baselines. 
\textbf{Fine-tuning (FT)}: training \(f_\theta\) with pre-trained weight \(\theta_\mathrm{s}\).
\textbf{FNP}~\cite{Hariharan_ICCV17_feature_penalty}: fine-tuning \(\ell_2\) penalty on \(g_\phi(x)\), i.e., \(\|g_\phi(x)\|^2_2\).
\textbf{L2SP}~\cite{xuhong_ICML18_l2sp}: fine-tuning \(f_\theta\) with an \(\ell_2\) penalty on \(\theta\) not to diverge from \(\theta_\mathrm{s}\), i.e., \(\|\theta-\theta_\mathrm{s}\|^2_2\).
\textbf{DELTA}~\cite{li_ICLR19_delta}: fine-tuning \(f_\theta\) with a penalty on the intermediate output not to diverge from one of \(\theta_\mathrm{s}\).
\textbf{BSS}~\cite{Chen_NeurIPS19_BSS}: fine-tuning \(f_\theta\) by penalizing the singular values of the batch feature matrices.
\textbf{RandReg}~\cite{Zhong_CVPR20_RandReg}: fine-tuning \(f_\theta\) with a penalty term to minimize the gap between feature vectors \(g_\phi(x)\) and reference vectors \(z\sim p(z)\), i.e., \(\|g_\phi(x) - z\|^2_2\).
\textbf{Core-tuning}~\cite{Zhang_NeurIPS21_coretuning}: fine-tuning \(f_\theta\) with supervised focal contrastive loss.
\textbf{DR-Tune}~\cite{Zhou_ICCV23_drtune}: fine-tuning \(f_\theta\) by penalizing \(W\) to classify the features extracted from source models into target classes.
\vspace{-3mm}
\paragraph{Baselines w/ source information.}
To assess the practicality of AdaRand, we additionally used the following baseline methods requiring auxiliary source information.
\textbf{Co-Tuning}~\cite{you_NeurIPS20_co_tuning}: fine-tuning \(f_\theta\) with simultaneous pseudo source tasks defined by mapping target and source class labels.
\textbf{UOT}~\cite{Liu_NeurIPS22_UOT}: fine-tuning \(f_\theta\) with simultaneous partial source tasks by extracting source samples related to target tasks with optimal-transport-based mapping algorithm.
\vspace{-3mm}

\begin{table*}[t]
    \centering
    \caption{Top-1 test accuracy (\%) of various combinations of pre-training methods and neural network architectures (Cars).}
    \label{tb:ex_pre_method_arch}
            \resizebox{0.9\textwidth}{!}{
        \begin{tabular}{lc|ccccc}\toprule
            Pre-training Method & Architecture & Fine-tuning & FNP~\cite{Hariharan_ICCV17_feature_penalty} &  DR-Tune~\cite{Zhou_ICCV23_drtune} & RandReg-Best & AdaRand (Ours) \\
          \midrule
            ImageNet Classification & RN-50 &  89.14\(^{\pm\text{.42}}\) & 90.27\(^{\pm\text{.10}}\) & 90.38\(^{\pm\text{.59}}\) & 90.61\(^{\pm.24}\) & \textbf{91.17}\(^{\pm\textbf{.13}}\) \\
            ImageNet Classification & ViT-B/32 &  78.56\(^{\pm\text{1.3}}\) & 81.77\(^{\pm\text{.21}}\) & 79.49\(^{\pm\text{.51}}\) & 82.46\(^{\pm{.20}}\) & \textbf{83.84}\(^{\pm\textbf{.13}}\) \\
            ImageNet Classification & ViT-B/16 &  87.35\(^{\pm\text{.53}}\) & 88.75\(^{\pm\text{.44}}\) & 88.19\(^{\pm\text{.26}}\) & 88.88\(^{\pm{.26}}\) & \textbf{89.54}\(^{\pm\textbf{.17}}\) \\
            ImageNet SimCLR~\cite{Chen_ICML20_SimCLR}   & RN-50 &  83.73\(^{\pm\text{.73}}\) & 84.53\(^{\pm\text{.32}}\) & 84.05\(^{\pm\text{.17}}\) &  84.08\(^{\pm.21}\) & \textbf{85.51}\(^{\pm\textbf{.05}}\) \\
            ImageNet Barlow Twins~\cite{Zbontar_ICML21_barlowtwins} & RN-50&  86.98\(^{\pm\text{.16}}\) & 87.44\(^{\pm\text{.15}}\) & 86.69\(^{\pm\text{.23}}\) &  87.74\(^{\pm.33}\) & \textbf{88.23}\(^{\pm\textbf{.39}}\) \\
            CLIP~\cite{Radford_ICML21_CLIP}   & RN-50 & 88.72\(^{\pm\text{.24}}\) & 90.19\(^{\pm\text{.40}}\) & 90.16\(^{\pm\text{.22}}\) & 90.78\(^{\pm.07}\) & \textbf{91.25}\(^{\pm\textbf{.63}}\) \\
            CLIP~\cite{Radford_ICML21_CLIP}   & ViT-B/32 &  83.56\(^{\pm\text{.60}}\) & 85.79\(^{\pm\text{.52}}\)& 85.71\(^{\pm\text{.59}}\)& 86.83\(^{\pm.34}\) & \textbf{87.40}\(^{\pm\textbf{.48}}\) \\
            CLIP~\cite{Radford_ICML21_CLIP}   & ViT-B/16 &  90.35\(^{\pm\text{.23}}\) & 91.24\(^{\pm\text{.03}}\) & 90.47\(^{\pm\text{.26}}\) & 91.33\(^{\pm.44}\) & \textbf{92.84}\(^{\pm\textbf{.48}}\) \\
            \bottomrule
        \end{tabular}
        }
\end{table*}

\begin{table}[t]
    \centering
    \caption{Top-1 test accuracy (\%) on multiple datasets (ResNet-50 pre-trained with ImageNet classification).}
    \label{tb:ex_multi_dataset}
        \resizebox{1.0\columnwidth}{!}{
        \begin{tabular}{lccccccccc}\toprule
            Method / Dataset & Aircraft & Birds & Cars & DTD & Flower & Pets  \\
          \midrule
            Fine-tuning    &  67.78\(^{\pm\text{.09}}\) & 74.80\(^{\pm\text{.56}}\) &  88.29\(^{\pm\text{.12}}\) & 63.88\(^{\pm\text{.31}}\) & 94.58\(^{\pm\text{.21}}\) & 89.45\(^{\pm\text{.18}}\) \\
            FNP~\cite{Hariharan_ICCV17_feature_penalty}    & 70.27\(^{\pm\text{.42}}\) & 79.57\(^{\pm\text{.41}}\) &  90.27\(^{\pm\text{.10}}\) & 72.13\(^{\pm\text{.04}}\) & 95.07\(^{\pm\text{.06}}\) & 91.21\(^{\pm\text{.12}}\) \\
            DR-Tune~\cite{Zhou_ICCV23_drtune}    &  73.89\(^{\pm\text{.10}}\) & 76.63\(^{\pm\text{.31}}\) &  90.38\(^{\pm\text{.59}}\) & 70.55\(^{\pm\text{.28}}\) & 93.43\(^{\pm\text{.17}}\) & \textbf{93.24}\(^{\pm\textbf{.11}}\) \\
            RandReg-\(U(0,1)\)  &  71.65\(^{\pm\text{.59}}\) & 79.82\(^{\pm\text{.21}}\) &  90.42\(^{\pm\text{.24}}\) & 72.32\(^{\pm\text{.41}}\) & 96.02\(^{\pm\text{.05}}\) & 91.72\(^{\pm\text{.11}}\) \\
            RandReg-\(\mathcal{N}(0,1)\)    &  73.11\(^{\pm\text{.88}}\) & 80.48\(^{\pm\text{.06}}\) &  90.32\(^{\pm\text{.41}}\) & 72.57\(^{\pm\text{.26}}\) & 95.01\(^{\pm\text{.11}}\) & 91.80\(^{\pm\text{.12}}\) \\
            RandReg-\(\mathcal{N}(\mu_\mathrm{s},\sigma^2_\mathrm{s})\) & 72.08\(^{\pm\text{.50}}\) & 80.20\(^{\pm\text{.01}}\) & 90.61\(^{\pm\text{.24}}\) & 71.77\(^{\pm\text{.78}}\) & 95.16\(^{\pm\text{.15}}\) & 91.81\(^{\pm\text{.11}}\) \\
            RandReg-CP & 72.10\(^{\pm\text{.20}}\) & 80.02\(^{\pm\text{.20}}\) & 90.55 \(^{\pm\text{.17}}\) & 72.34\(^{\pm\text{.27}}\) & 95.86\(^{\pm\text{.30}}\) & 91.76\(^{\pm\text{.61}}\)\\
            AdaRand (Ours) &  {\bf 74.60}\(^{\pm\textbf{.10}}\) & {\bf 81.27}\(^{\pm\textbf{.26}}\) &  {\bf 91.17}\(^{\pm\text{.13}}\) & {\bf 74.86}\(^{\pm\textbf{.22}}\) & {\bf 96.68}\(^{\pm\textbf{.14}}\) & 92.34\(^{\pm\text{.26}}\) \\
            \bottomrule
        \end{tabular}
        }
\end{table}

\paragraph{Datasets.}
We used six image datasets for classification tasks in various domains: \textbf{Aircraft}~\cite{maji_13_aircraft}, \textbf{Birds}~\cite{Welinder_10_cub2002011}, \textbf{Cars}~\cite{krause_3DRR2013_stanford_cars}, \textbf{DTD}~\cite{cimpoi_CVPR14_DTD}, \textbf{Flowers}~\cite{Nilsback_08_flowers}, and \textbf{Pets}~\cite{parkhi_CVPR12_oxford_pets}.
Furthermore, we reduced the Cars dataset by \(\{10,25,50\}\%\) in volume on a fixed random seed to evaluate smaller dataset cases.
We randomly split a dataset into \(9:1\) and used the former as the training set and the latter as the validation set.
\vspace{-3mm}
\paragraph{Architectures.}
We basically used the \textbf{ResNet-50} architecture~\cite{he_resnet}.
To confirm the flexibility among architectures, we also evaluated our method with \textbf{ViT-B}~\cite{Dosovitskiy_ICLR21_ViT}.
\vspace{-3mm}
\paragraph{Training.}
As the pre-training methods, we used supervised classification~\cite{yosinski_NeurIPS14_transferable} and self-supervised pre-training (SimCLR~\cite{Chen_ICML20_SimCLR} and Barlow Twins~\cite{Zbontar_ICML21_barlowtwins}) on ImageNet.
Further, we used the pre-trained weights of CLIP~\cite{Radford_ICML21_CLIP}.
For training on downstream classification tasks with ResNet-50, we trained \(f_\theta\) by the Nesterov momentum SGD for 200 epochs with a momentum of 0.9, and an initial learning rate of 0.01; we decayed the learning rate by 0.1 at 60, 120, and 160 epochs.
For ViT-B, we used the AdamW~\cite{Loshchilov_ICLR19_AdamW} optimizer with the initial learning rate of 3.0\(\times\)10\(^{-5}\) that decayed by cosine annealing.
We used mini-batch sizes of 64.
The input samples were resized into a resolution of \(224\times224\). 
We used \(\lambda\) of \(1.0\); we discuss the effect of \(\lambda\) in the supplement.
We selected the final model by checking the validation accuracy for each epoch.
We implemented the training and evaluation with PyTorch-1.11.
We ran the experiments three times on a 24-core Intel Xeon CPU with an NVIDIA A100 GPU with 40GB VRAM and recorded average test accuracy with standard deviation evaluated on the final models.

\subsection{Evaluation on Multiple Pre-trained Models}\label{sec:ex_multi_arch}
One of our motivations is to develop a method that achieves high performance on arbitrary pre-training methods without relying on auxiliary source information and additional heavy computation.
Here, to verify that this motivational goal is achieved, we evaluate AdaRand with multiple combinations of pre-training methods and architectures such as CLIP and ViT-B.
In this experiment, we compare our method with RandReg, FNP, and DR-Tune, which are available without auxiliary source information.
Table~\ref{tb:ex_pre_method_arch} shows that AdaRand stably outperforms RandReg and the other baselines for all combinations of pre-training methods and architectures.
In particular, AdaRand overcomes the negative effects on SimCLR observed in Sec.~\ref{sec:analisys}.
These results indicate that the adaptive prior update of AdaRand is widely effective for many pre-trained models.

\begin{table}[t]
    \centering
    \caption{Top-1 test accuracy (\%) on small training datasets (Cars, ResNet-50 pre-trained with ImageNet classification).}
    \label{tb:ex_small_data}
        \resizebox{1.0\columnwidth}{!}{
        \begin{tabular}{lccc}\toprule
            Method / Dataset Size (\%) & 10 \% & 25 \% & 50\%  \\
          \midrule
            Fine-tuning    &  19.58\(^{\pm\text{.07}}\) & 53.10\(^{\pm\text{.08}}\) &  77.86\(^{\pm\text{.09}}\) \\
            FNP~\cite{Hariharan_ICCV17_feature_penalty}    &  23.68\(^{\pm\text{.34}}\) & 57.07\(^{\pm\text{.34}}\) &  79.93\(^{\pm\text{.21}}\) \\
            DR-Tune~\cite{Zhou_ICCV23_drtune}    &  23.68\(^{\pm\text{.34}}\) & 57.07\(^{\pm\text{.34}}\) &  79.93\(^{\pm\text{.21}}\) \\
            RandReg-\(U(0,1)\)  &  25.35\(^{\pm\text{.14}}\) & 58.33\(^{\pm\text{.45}}\) &  80.91\(^{\pm\text{.51}}\) \\
            RandReg-\(\mathcal{N}(0,1)\)    &  26.15\(^{\pm\text{.21}}\) & 59.94\(^{\pm\text{.58}}\) &  81.74\(^{\pm\text{.41}}\) \\
            RandReg-\(\mathcal{N}(\mu_0, \sigma^2_0)\)    &  24.95\(^{\pm\text{.05}}\) & 59.66\(^{\pm\text{.24}}\) &  81.31\(^{\pm\text{.17}}\)  \\
            RandReg-CP    &  26.45\(^{\pm\text{.26}}\) & 59.37\(^{\pm\text{.46}}\) &  81.13\(^{\pm\text{.23}}\)  \\
            AdaRand (Ours) &  {\bf 27.55}\(^{\pm\textbf{.10}}\) & {\bf 61.09}\(^{\pm\textbf{.54}}\) &  {\bf 82.19}\(^{\pm\text{.19}}\) \\
            \bottomrule
        \end{tabular}
        }
\end{table}

\begin{table}[t]
    \centering
    \caption{Comparison among fine-tuning regularization methods (Cars, ResNet-50 pre-trained with ImageNet classification). We measure top-1 test accuracy, averaged training time per epoch, and GPU memory usage. Our method (AdaRand) outperforms the other baselines in accuracy while maintaining reasonable computation time and GPU memory consumption.}
    \label{tb:ex_ft_methods}
        \resizebox{1.0\columnwidth}{!}{
        \begin{tabular}{lccc}\toprule
            Method & Test Accuracy (\%) & Time / Epoch (sec.) & GPU Mem. (MiB)\\
          \midrule
            Fine-tuning & 88.29\(^{\pm\text{.12}}\) & \textbf{9.737} & \textbf{8,073} \\
            FNP~\cite{Hariharan_ICCV17_feature_penalty}  & 90.27\(^{\pm\text{.06}}\) & 9.916 & 8,073 \\
            L2SP~\cite{xuhong_ICML18_l2sp} & 88.50\(^{\pm\text{.25}}\) & 12.405 & 8,153 \\
            DELTA~\cite{li_ICLR19_delta}  & 89.00\(^{\pm\text{.24}}\) & 14.733 & 9,211\\
            BSS~\cite{Chen_NeurIPS19_BSS}  & 89.70\(^{\pm\text{.06}}\) & 11.403 & 8,227 \\
            Core-tuning~\cite{Zhang_NeurIPS21_coretuning}  & 90.01\(^{\pm\text{.13}}\) & 16.872 & 24,223 \\
            DR-Tune~\cite{Zhou_ICCV23_drtune} & 90.38\(^{\pm\text{.59}}\) & 25.475 & 8,845 \\\midrule
            Co-Tuning~\cite{you_NeurIPS20_co_tuning} & 90.66\(^{\pm\text{.34}}\) & 11.546 & 8,125 \\
            UOT~\cite{Liu_NeurIPS22_UOT} & 90.82\(^{\pm\text{.19}}\) & 12.405 & 14,293 \\\midrule
            RandReg-\(U(0,1)\) & 90.42\(^{\pm\text{.24}}\) & 9.988 & 8,075 \\
            RandReg-\(\mathcal{N}(0,1)\) & 90.32\(^{\pm\text{.41}}\) & 9.971 & 8,075 \\
            RandReg-\(\mathcal{N}(\mu_\mathrm{s},\sigma^2_\mathrm{s})\) & 90.61\(^{\pm\text{.24}}\) & 9.963 & 8,075 \\
            RandReg-CP & 90.55\(^{\pm\text{.17}}\) & 11.552 & 8,075\\
            AdaRand w/o \(\ell_\mathrm{intra}\)  & 90.81\(^{\pm\text{.21}}\) & 12.410 & 8,585\\
            AdaRand w/o \(\ell_\mathrm{inter}\)  & 90.99\(^{\pm\text{.06}}\) & 12.841 & 8,585\\
            AdaRand (Ours)  & \textbf{91.17}\(^{\pm\textbf{.13}}\) & 13.824 & 8,585\\
            \bottomrule
        \end{tabular}
        }
\end{table}

\begin{figure*}[t]
    \centering
        \begin{tabular}{ccc}
          \begin{minipage}[t]{0.3\textwidth}
                \includegraphics[align=c,width=5cm]{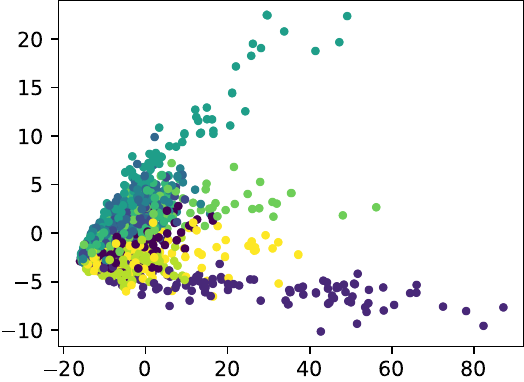}\subcaption{Fine-tuning} 
          \end{minipage}&
          \begin{minipage}[t]{0.3\textwidth}
                \includegraphics[align=c,width=5cm]{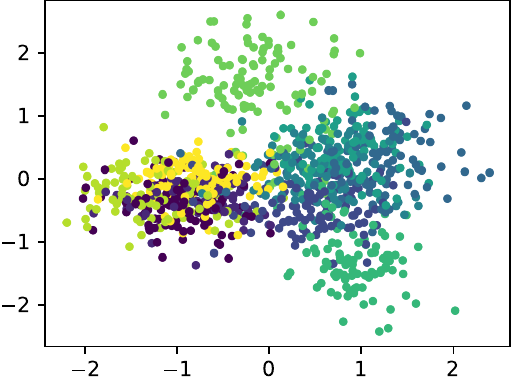}\subcaption{RandReg-\(U(0,1)\)}
          \end{minipage}&
          \begin{minipage}[t]{0.3\textwidth}
                \includegraphics[align=c,width=5cm]{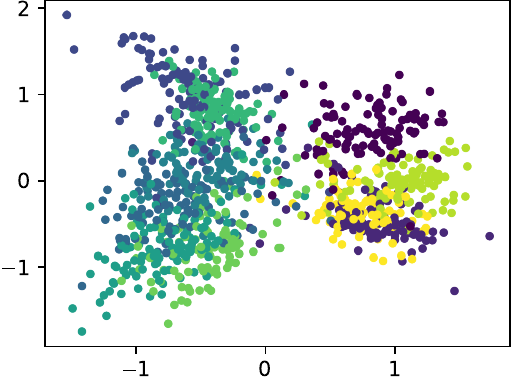}\subcaption{RandReg-\(\mathcal{N}(0,1)\)}
          \end{minipage}\\
          \begin{minipage}[t]{0.3\textwidth}
                \includegraphics[align=c,width=5cm]{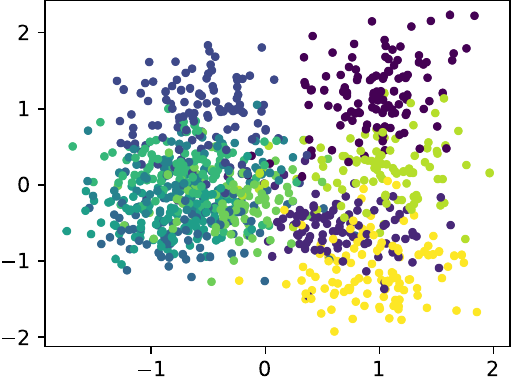}\subcaption{RandReg-\(\mathcal{N}(\mu_\mathrm{s},\sigma^2_\mathrm{s})\)} 
          \end{minipage}&
          \begin{minipage}[t]{0.3\textwidth}
                \includegraphics[align=c,width=5cm]{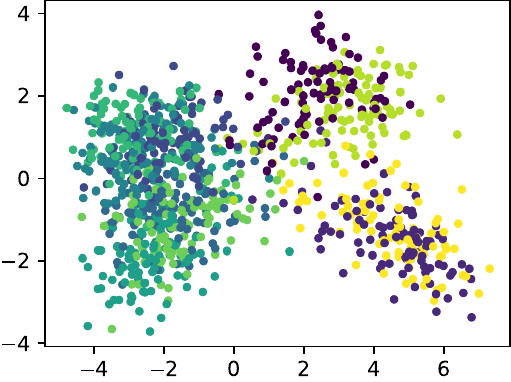}\subcaption{RandReg-CP}
          \end{minipage}&
          \begin{minipage}[t]{0.3\textwidth}
                \includegraphics[align=c,width=5cm]{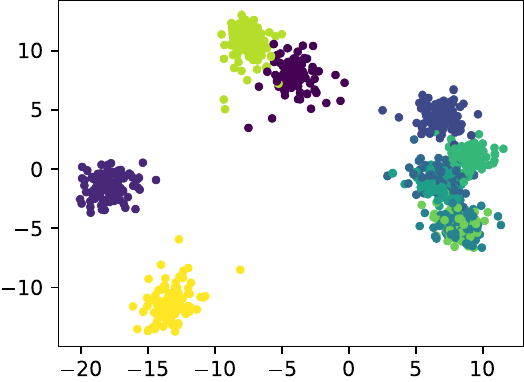}\subcaption{AdaRand (Ours)}
          \end{minipage}\\
        \end{tabular}
        \caption{
            PCA visualization of feature spaces of trained models (CIFAR-10, ResNet-50).
            The colors in the sample plot correspond to that class.
            AdaRand clearly forms well-separated clusters, which can be useful for solving downstream classification tasks.
            }
        \label{fig:feature_viz}
        \vspace{-3mm}
 \end{figure*}

\subsection{Evaluation on Multiple Datasets}\label{sec:ex_multi_datasets}
We evaluate AdaRand on multiple different datasets to demonstrate its generality across the datasets.
Table~\ref{tb:ex_multi_dataset} shows the results.
Except for the Pets dataset, AdaRand achieved the best results for all of the datasets.
While the performance of RandReg depends on the hyperparameters of priors, AdaRand stably performs by the adaptive prior update.

\subsection{Evaluation on Small Datasets}\label{sec:ex_small_datasets}
We evaluate AdaRand by reducing the data volume of Cars into \(\{10, 25, 50\}\)\%.
Table~\ref{tb:ex_small_data} shows that AdaRand still outperforms the baselines even with a limited dataset of less than a few training samples per class (i.e., 10\%).
The results demonstrate that AdaRand performs well in data-scarce scenarios.

\subsection{Evaluation on Comparison to SoTA Methods}\label{sec:ex_sota}
We demonstrate the efficacy of AdaRand by comparing it with the SoTA baselines.
Table~\ref{tb:ex_ft_methods} shows the results on the Cars dataset with the ResNet-50 architecture pre-trained on the ImageNet classification task.
We confirm that our AdaRand achieved better performance than the methods leveraging auxiliary source information (i.e., Co-Tuning and UOT) or additional heavy computation costs (i.e., Core-tuning and DR-Tune).
The bottom part of Table~\ref{tb:ex_ft_methods} also shows an ablation study of AdaRand, where RandReg-CP is a method using the conditional prior defined in Eq.~(\ref{eq:adarand_prior}) without updating them.
AdaRand significantly improved the performance of RandReg-CP and the losses of \(\ell_\mathrm{inter}\) and \(\ell_\mathrm{intra}\) complementarily contributed to the improvements.
This indicates that na\"ively introducing the conditional prior is not sufficient to solve the challenges of RandReg and adaptively updating priors is effective in terms of performance.

\subsection{Analysis and Discussion}\label{sec:ex_analysis}
Through the experiments in the previous sections, we validate that AdaRand stably improves the accuracy of the downstream classification tasks in various settings and overcomes the RandReg's challenge caused by using fixed priors discussed in the former paragraph of Sec.~\ref{sec:pre_challenges}.
Here, we provide the analysis of AdaRand to confirm whether it overcomes the rest challenges of RandReg, i.e., decreasing feature norms \(\|g_\phi(x)\|^2_2\) and entropy \(H(g_\phi(x))\).
Figure~\ref{fig:norm_entropy} demonstrates that AdaRand succeeds in preventing the decreasing \(\|g_\phi(x)\|^2_2\) and \(H(g_\phi(x))\).
As a result, AdaRand achieves better the gradient norm of cross-entropy loss \(\|\nabla_W \ell_\mathrm{CE}(f_\theta(x), y)\|^2_2\) and mutual information \(I(g_\phi(x);y)\) than RandReg.
This also can be an explanation for the reason why AdaRand stably outperforms RandReg in the previous sections.

Finally, we discuss the effects of AdaRand on feature spaces by visualizing feature vectors from trained models with the PCA dimension reduction.
As the dataset, we used CIFAR-10~\cite{krizhevsky09_cifar10}.
We trained ResNet-50 by each method and reduced the dimensions of the output of \(g_\phi(x)\) by PCA.
We randomly selected 1,024 samples from the test set for the input.
The visualization results are illustrated in Figure~\ref{fig:feature_viz}.
While the baseline method forms less separated feature clusters, AdaRand clearly forms independent and dense feature clusters for each class.
This indicates that the adaptive prior update of AdaRand helps models learn useful representations for solving downstream classification tasks. 

\section{Conclusion}\label{sec:conclusion}
This paper presents a novel regularization method for fine-tuning deep neural networks called AdaRand.
AdaRand penalizes feature vectors of deep models by guiding to the random reference vectors that are generated from the class conditional prior distributions.
To encourage the fine-tuning models to generate useful features for target classification tasks, AdaRand adaptively updates the conditional prior distributions so that the prior distributions are close to the current feature distribution and balance the distance between classes.
Through this simple method, The fine-tuned model increases the amount of mutual information between features and labels, resulting in a significant improvement in final test accuracy without either auxiliary source information or additional heavy computation costs.
An important future step is to extend AdaRand beyond discriminative tasks such as generative modeling by diffusion models.

\clearpage
{\small
\bibliographystyle{ieeenat_fullname}
\bibliography{ref}
}

\clearpage

\onecolumn
\appendix
\renewcommand\thetable{\Roman{table}}
\renewcommand\thefigure{\Roman{figure}}

\begin{figure*}[h]
    \centering
    \begin{minipage}{0.49\linewidth}
    \centering
    \includegraphics[width=0.8\columnwidth]{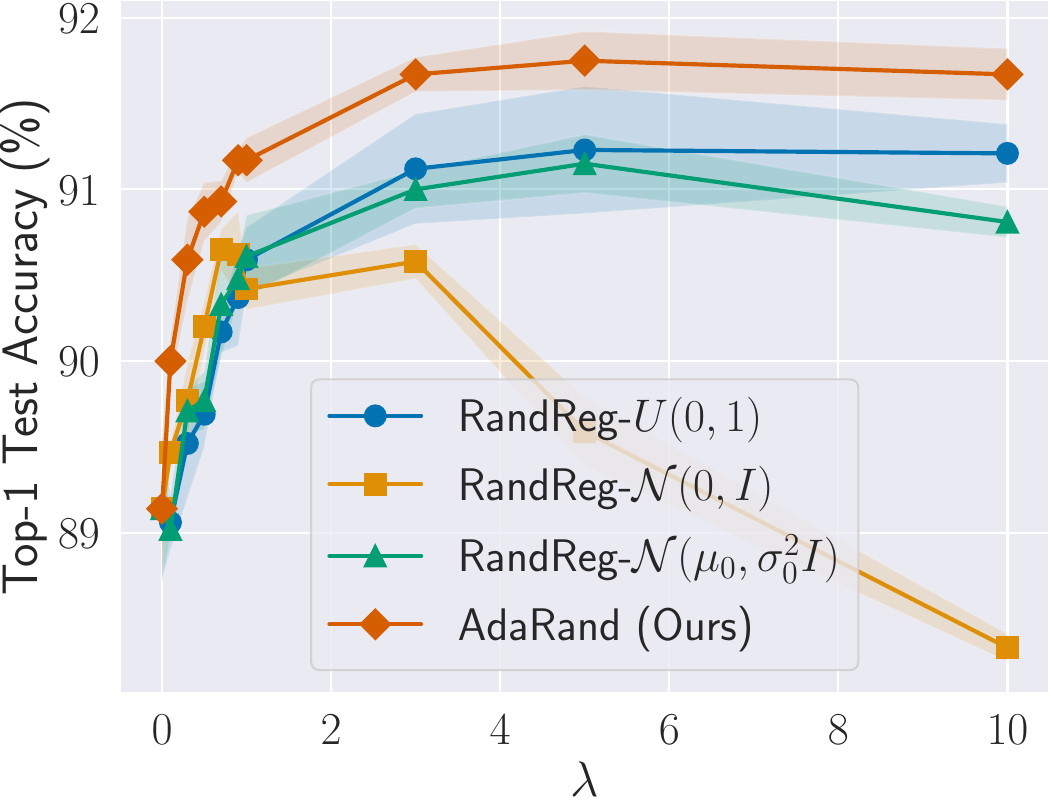}
    \caption{
    Top-1 Accuracy on Cars (ResNet-50)
    }
    \label{fig:lambda}
    \end{minipage}
    \begin{minipage}{0.49\linewidth}
    \centering
    \captionof{table}{Top-1 Test Accuracy on Cars (ResNet-50)}
    \label{tb:ex_ema_alpha}
        \begin{tabular}{lc}\toprule
            EMA decay param. \(\alpha\) & Test Accuracy (\%) \\
          \midrule
            0.0 (Only using \(\hat{\mu}_k\)) & 90.89\(^{\pm\text{.10}}\)\\
            0.1 & 90.96\(^{\pm\text{.20}}\)\\
            0.3 & 91.14\(^{\pm\text{.26}}\)\\
            0.5 & 91.17\(^{\pm\text{.13}}\)\\
            0.7 & 91.32\(^{\pm\text{.04}}\)\\
            0.9 & 91.27\(^{\pm\text{.12}}\)\\
            0.99 & 91.13\(^{\pm\text{.18}}\)\\
            0.999 & 91.26\(^{\pm\text{.19}}\)\\
            1.0 (Fixed \(\bar{\mu}_k\)) & 90.62\(^{\pm\text{.04}}\)\\
            \bottomrule
        \end{tabular}
    \end{minipage}
\end{figure*}

\begin{figure*}[t]
    \centering
    \begin{minipage}{0.45\linewidth}
    \caption{\footnotesize Fine-tuning CLIP (ViT-B/16).}
    \label{tb:clip_in}
        \vspace{-2mm}
        \resizebox{1.0\columnwidth}{!}{
        \begin{tabular}{lccc}\toprule
            Method & ImageNet & ImageNet-A & ImageNet-R \\
          \midrule
            Fine-tuning    &  79.74  & 18.31 & 45.40\\
            FNP    &  79.95 & 20.36 & 48.31 \\
            DR-Tune    &  31.91  & 1.51 & 4.48\\
            RandReg-\(U(0,1)\)  &  80.08 & 20.27 & 47.01 \\
            RandReg-\(\mathcal{N}(0,1)\)    &  77.35 & 17.47 & 43.25 \\
            RandReg-\(\mathcal{N}(\mu_0, \sigma^2_0)\)  & 80.04 & 20.27 & 47.79 \\
            AdaRand (Ours) & \textbf{82.00} & \textbf{23.50} & \textbf{51.72}\\
            \bottomrule
        \end{tabular}
        }
        \end{minipage}
    \begin{minipage}{0.45\linewidth}
        \caption{\footnotesize Object Detection and Segmentation on VOC-2012.}
        \label{tb:detseg}
        \vspace{-2mm}
        \resizebox{1.0\columnwidth}{!}{
        \begin{tabular}{lcc}\toprule
            Method & bbox AP & mask AP \\
          \midrule
            Fine-tuning    &  43.5 & 32.2 \\
            RandReg-\(U(0,1)\)  &  42.8 & 31.3\\
            RandReg-\(\mathcal{N}(0,1)\) & 41.2 & 30.1\\
            AdaRand (Ours) &  {\bf 45.9} & {\bf 35.2} \\
            \bottomrule
        \end{tabular}
        }
        \end{minipage}
\end{figure*}

\section{Effects of Hyperparameters}\label{sec:effect_hyperparam}
In the main paper, we fixed the hyperparameters of the trade-off parameter \(\lambda\) in Eq.~(1) and decay parameter \(\alpha\) in Eq.~(14) for fair comparison.
Here, we confirm the effects of varying them on the performance.

\paragraph{Trade-off parameter \(\lambda\) in Eq. (1).}
The trade-off parameter \(\lambda\) in Eq. (1) controls the regularization effect by the random reference vectors. We fixed \(\lambda=1.0\) for a fair comparison, but evaluating the sensitivity of \(\lambda\) is important in practice.
Figure~\ref{fig:lambda} shows the results varying \(\lambda\) among \(\{0.0, 0.1, 0.3, 0.5, 0.7, 0.9, 1.0, 3.0, 5.0, 10.0\}\).
We observed that our AdaRand stably outperformed the RandReg variants by large margins.
This indicates that AdaRand is not sensitive to the choice of \(\lambda\).

\paragraph{EMA decay parameter $\alpha$ in Eq. (14).}
As described in Sec.~5.2 in the main paper, we update the prior parameter set \(\boldsymbol{\mu}\) according to the running mean vector set \(\boldsymbol{\bar{\mu}}\).
To compute \(\boldsymbol{\bar{\mu}}\), we introduce the exponential moving average in Eq.~(14) and (15) as
\begin{eqnarray}
    &\bar{\mu}_k \leftarrow \alpha \bar{\mu}_k + (1-\alpha) \hat{\mu}_k,\nonumber\\
    &\hat{\mu}_k = \frac{1}{B_k}\sum^{B_k}_{i=1} g_\phi(x^i),\nonumber
\end{eqnarray}
That is, the hyperparameter \(\alpha\) controls the decay of the past information in \(\boldsymbol{\bar{\mu}}\).
If \(\alpha\) is zero, then we use only batch-wise feature mean vectors for updating \(\boldsymbol{\mu}\), and if \(\alpha\) is one, then we fix \(\boldsymbol{\bar{\mu}}\) in its initial state.
We evaluated the effect of \(\alpha\) on the Cars dataset.
Table~\ref{tb:ex_ema_alpha} shows that the performance of AdaRand has certain robustness to the choice of \(\alpha\) as long as it uses the EMA update of \(\boldsymbol{\bar{\mu}}\), i.e., \(0 < \alpha < 1\), and setting \(\alpha\) to the range of \([0.5,0.9]\) achieves better accuracy scores than the others.
This indicates that updating \(\boldsymbol{\bar{\mu}}\) by EMA is important to the performance improvements, and we can slightly improve the performance by tuning \(\alpha\).

\section{Additional Experiments}\label{sec:additional_experiments}

\subsection{Fine-tuning CLIP on ImageNet}
We evaluate our method on large-scale target datasets.
Table~\ref{tb:clip_in} shows the result of fine-tuning CLIP on ImageNet (5 epoch).
Our method is effective even when the class numbers and training dataset are large, while RandReg sometimes degrades baselines.
In particular, our method significantly improves the performance of ImageNet-R, indicating that our method also enhances the generalizability to real-world distribution shifts.

\subsection{Application to Object Detection}
We evaluate our method on non-classification tasks to show the generalizability across tasks.
We tested our method on the Pascal VOC-2012 object detection and semantic segmentation with Mask-RCNN.
Table~\ref{tb:detseg} shows that our method boosts the baseline even on non-classification tasks.
Note that our method can be na\"ively applied to the Pascal VOC-2012 because each image has a single category object in the dataset, but it is non-trivial when input images have multiple category objects like the COCO dataset.
We will explore extending our method to arbitrary tasks in future work.

\end{document}